\documentclass[a4paper,conference]{IEEEtran}
\IEEEoverridecommandlockouts

\usepackage{cite}
\usepackage{amsmath,amssymb,amsfonts}
\usepackage{algorithmic}
\usepackage{graphicx}
\usepackage{textcomp}
\usepackage{xcolor}
\usepackage{url}
\usepackage{stfloats}
\usepackage{multicol,multirow}
\usepackage{url}
\usepackage[normalem]{ulem}
\usepackage{hyperref}
\usepackage{cleveref}
\usepackage{authblk}
\usepackage{booktabs}

\def\BibTeX{{\rm B\kern-.05em{\sc i\kern-.025em b}\kern-.08em
    T\kern-.1667em\lower.7ex\hbox{E}\kern-.125emX}}
\begin{document}

\title{HREB-CRF: Hierarchical Reduced-bias EMA for Chinese Named Entity Recognition}

\author{
\textbf{Sijin Sun}$^{1,4,\dagger}$ \hspace{10pt}
\textbf{Ming Deng}$^{2,\dagger}$ \hspace{10pt}
\textbf{Xingrui Yu}$^{1,3}$ \hspace{10pt}
\textbf{Liangbin Zhao}$^{1,*}$
}
\affil{\normalsize
$^\dagger$Equal contribution \quad $^*$Corresponding author \\
$^1$ Institute of High Performance Computing, Agency for Science, Technology and Research (A*STAR IHPC), Singapore \\
$^2$ Shanghai University \\ 
$^3$ Centre for Frontier AI Research, Agency for Science, Technology and Research (A*STAR CFAR), Singapore \\
$^4$ National University of Singapore
}

\maketitle

\begin{abstract}
Incorrect boundary division, complex semantic representation, and differences in pronunciation and meaning often lead to errors in Chinese Named Entity Recognition (CNER). To address these issues, this paper proposes HREB-CRF framework: Hierarchical Reduced-bias EMA with CRF. The proposed method amplifies word boundaries and pools long text gradients through exponentially fixed-bias weighted average of local and global hierarchical attention. Experimental results on the MSRA, Resume, and Weibo datasets show excellent in F1, outperforming the baseline model by 1.1\%, 1.6\%, and 9.8\%. The significant improvement in F1 shows evidences of strong effectiveness and robustness of approach in CNER tasks. Code available at \url{https://github.com/StanleySun233/HREB-CRF}.

\end{abstract}


\section{Introduction}
\label{sec:introduction}
Token classification is one of the key research directions in NLP, where NER assigns entity types to each input token for extracting critical information. An effective NER system not only bolsters tasks such as sentiment analysis, compliance monitoring, and recommendation systems, but also enables deeper text comprehension and facilitates automated processing across diverse contexts. Classic approaches like Lattice-LSTM\cite{zhang2018chinese} laid the groundwork for structured Chinese NER, while subsequent methods, including Flat-Transformer\cite{li2020flat} and BERT-based solutions\cite{jia2020entity}, have demonstrated notable success.

However, CNER poses unique challenges due to the complexity of the language. 1) \textbf{Lack of Explicit Word Boundaries:} Unlike English, Chinese texts are written without whitespace separation, making entity boundary detection especially difficult. 2) \textbf{Complex Semantics and Polysemy:} The rich morphology, polysemy, and extended meanings of Chinese require precise context modeling to avoid confusion caused by homophones or nested entities. 3) \textbf{Segmentation Ambiguity:} Conventional text features (e.g., radicals, phonetics) often struggle with local segmentation errors and global dependencies, leading to suboptimal performance in real-world scenarios.

To address these issues, Exponential Moving Average (EMA)\cite{hunter1986exponentially} has garnered attention due to its ability to stabilize model updates and smooth out noisy signals, particularly beneficial for local boundary detection in time-series-like or sequential data. By exponentially decaying historical information, EMA can emphasize recent tokens and reduce the impact of earlier noise, which aligns well with the fine-grained boundary identification crucial in Chinese NER. Nevertheless, a simple EMA mechanism often underestimates long-range dependencies, which are vital for handling polysemy and nested structures in Chinese.

Therefore, inspired by the MEGA structure\cite{ma2022mega}, a Reduced-biased Hierarchical EMAttention framework (RHEMA) is proposed integrated with RoBERTa\cite{cui2020revisiting}. This hierarchical design, shown as \Cref{fig:overall-structure} ensures both short-range (local) and long-range (global) dependencies are jointly modeled, addressing segmentation ambiguities. The reduced-bias connection further mitigates the risk of gradient vanishing or exploding when stacking deep layers, thus providing more stable parameter updates.

The contributions of this work can be mainly summarized as the follows:

\begin{enumerate}
    \item The proposed work innovatively introduces RHEMA into the NER task, enabling more accurate modeling of short-range dependencies for entity boundary detection while preserving global contextual insights.

    \item The proposed work apply a reduced-bias method in each module to alleviate gradient vanishing, ensuring stable gradient flow and further enhancing the model’s ability to capture intricate linguistic features in deep architectures.
    
    \item By leveraging an advanced text embedding approach, the model achieves superior efficiency, stability, and generalization capabilities in text vectorization and complex language representation tasks, thereby significantly enhancing overall performance.
\end{enumerate}

\section{Related Work}
\label{sec:related_work}


Deep learning performs well in NER by capturing contextual information, learning abstract features automatically, and handling long-distance dependencies through recursive or convolutional operations. Integrating conditional random fields (CRF) further enhances performance by modeling label sequences globally, ensuring consistency and improving accuracy.

Recently development of these methods shows a more comprehensive understanding of contextual information and improves the generalization ability in multiple fields and multiple contexts.

Chinese NER faces unique challenges due to tokenization-induced semantic discrepancies \cite{qiu2020pre}. Character-based segmentation outperforms word-based approaches \cite{he2008chinese}, and while Word2Vec \cite{goldberg2014word2vec} struggles with polysemy, hybrid models like CWVF-BiLSTM-CRF improve accuracy by combining character and word inputs \cite{ye2020chinese}. Adversarial transfer learning aids boundary detection \cite{cao2018adversarial}, while Lattice-LSTM \cite{zhang2018chinese} and its variants (e.g., LR-CNN \cite{gui2019cnn}) leverage dictionaries and attention to resolve lexical conflicts. Multi-representation encoding \cite{liu2019encoding} and FLAT-Transformer’s lattice structures \cite{li2020flat} further advance performance.



Specialized data structures enhance NER feature capture. Tree-LSTM \cite{tai2015improved} integrates sub-features via topological networks, while DyLex \cite{wang2021dylex} augments BERT with updatable dictionaries. Liu et al.'s TFM \cite{liu2022tfm} improves character understanding through positional encoding. Besides, other works on exploiting Chinese linguistic features. Mai et al.\cite{mai2022pronounce} studies homophones, and Li et al. \cite{li2022dependency} uses syntactic dependency trees for global attention.

\section{Methodology}
\label{sec:methodology}

\subsection{Overview of HREB-CRF Architecture}

\begin{figure*}[!htbp]
    \centering
    \includegraphics[width=0.5\linewidth]{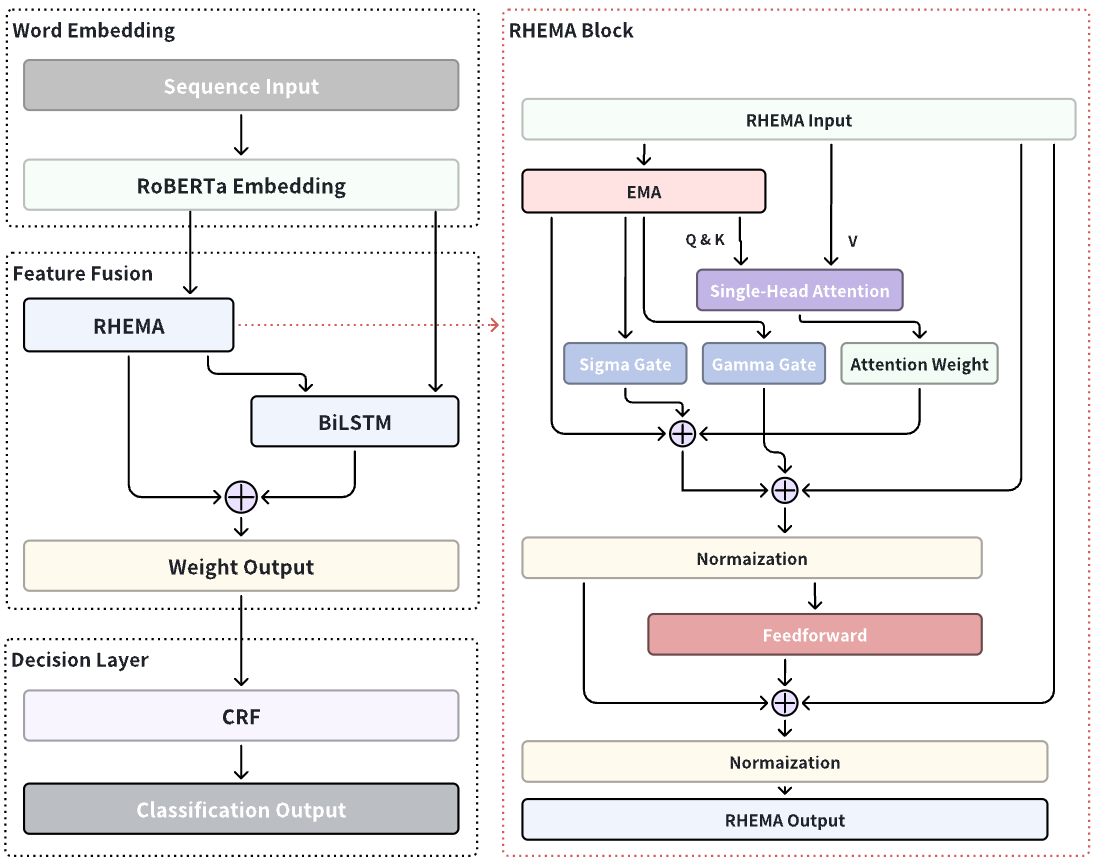}
    \caption{Overall Structure of proposed model. The left figure shows the end-to-end model architecture from text input to entity classification output, and the right figure shows the data processing flow of the RHEMA architecture.}
    \label{fig:overall-structure}
\end{figure*}

Proposed model is shown in \Cref{fig:overall-structure}. BERT structure model is used as the pre-training head, and output the start and end (B-Cls and last I-Cls structure annotation) features of the CRF-constrained entity. On this basis, the additional Hierarchical EMA module is used to enhance capturing the short-distance relationship of the text vector. Compared with traditional attention method, this improved method has the ability to extract both short-range and long-range features. Next, this hierarchical dependency weights are input into the Bi-LSTM layer to capture the context and strengthen the long-distance dependency capabilities, and finally input into the CRF for sequence dependency decoding. In addition, a reduced-biased module is added to each layer to reduce the possibility of gradient disappearance.

\subsection{Exponentially Moving Average}


Moving average method is widely used in sequence analysis tasks, such as stock and anomaly detection. It is proposed to reduce the global impact of local outliers anomalies. Generally speaking, most of the usage of MA are: simple moving average in \Cref{eq:sma}, weighted moving average in \Cref{eq:wma}, cumulative moving average in \Cref{eq:cma} and exponentially weighted moving average in \Cref{eq:ema}. 

\begin{equation}
    SMA_t = \frac{1}{N} \sum_{i=t-N+1}^{t} x_i
    \label{eq:sma}
\end{equation}

\begin{equation}
    WMA_t = \frac{\sum_{i=0}^{N-1} w_i \odot x_{t-i}}{\sum_{i=0}^{N-1} w_i}
    \label{eq:wma}
\end{equation}

\begin{equation}
    CMA_t = \frac{\sum_{i=1}^{t} x_i}{t}
\label{eq:cma}
\end{equation}

\begin{equation}
    EMA_t = \alpha \odot  x_t + (1 - \alpha) \odot  EMA_{t-1}
    \label{eq:ema}
\end{equation}




In the exponentially moving average (EM) formula, $\alpha$ is used as a weighting factor to determine the degree of attenuation of historical data. A larger $\alpha$ value indicates that current data is given more importance and the reliance on earlier data is weakened. Therefore, EM can effectively reduce the impact of random factors in the sequence on the overall trend, because as shift window goes by, $(1 - \alpha)^{t} \cdot y_{t}$ will gradually decrease and tend to zero, gradually diluting the weight of earlier data.

The advantage of EM lies in its high responsiveness to recent data, which can adapt to data trend changes more quickly while reducing lag. This feature is particularly suitable for capturing turning points and mutations in trends in sequences, especially for application scenarios such as time series data with turning points that require rapid weight adjustment. In the NER task, finding entity boundaries is one of the key issues. In addition, EM reduces the impact of long-term historical data through exponential decay, and can smooth noise without losing data continuity, making it robust in dynamic environments.

Similar to CNNs, EM can be regarded as applying additional decaying weights to each data block. In CNNs, the convolution kernel extracts local features through a sliding window, while in EMA, the weighting factor $\alpha$ plays a similar role to the convolution kernel, giving different importance to data of different time steps through gradually decaying weights. This decaying weight mechanism makes EM pay more attention to recent data and reduces the impact of early data on the overall mean. It is similar to the weight sharing and local perception characteristics in convolution operations. This feature gives EM sensitivity to changes in data trends, can smooth short-term fluctuations in the sequence, and retain information about longer trends, thereby playing a role in smoothing and denoising in the analysis and prediction of time series data.

\subsection{HEMA}



\begin{figure}
    \centering
    \includegraphics[width=0.6\linewidth]{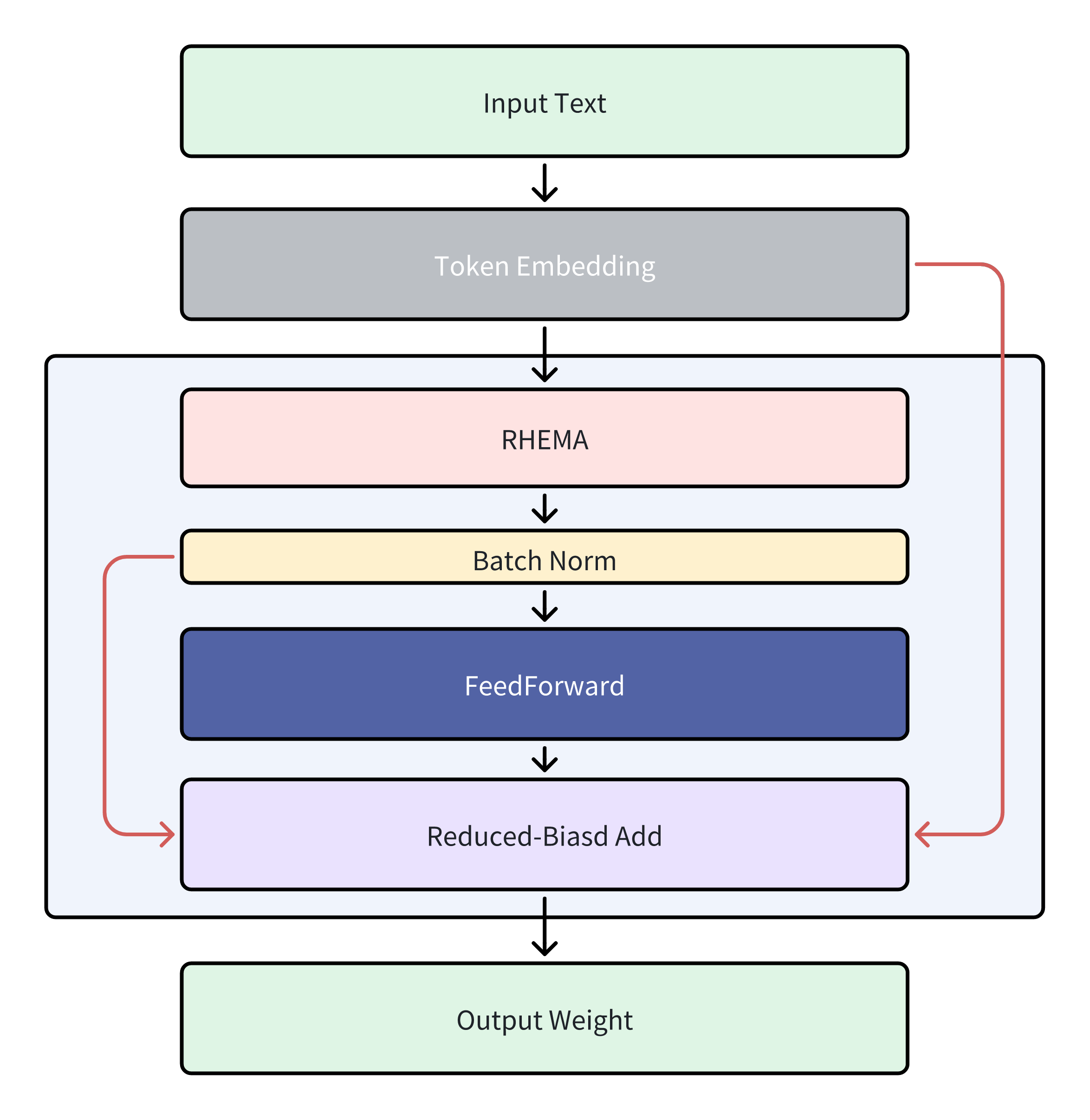}
    \caption{Proposed RHEMA architecture with added reduced-biased module. The figure illustrates the architecture of the RHEMA framework for text processing, enhanced by the addition of a reduced-bias module, marked in red. The core RHEMA block includes sequential processing through Batch Normalization and a FeedForward network. A novel "Reduced-Bias Add" module, shown with red arrows, is integrated to mitigate potential bias in the output, while preserving a feedback loop to earlier stages.}
    \label{fig:rhema-structure}
\end{figure}

Since the exponential averaging method is unique in the decay weight mechanism, we add the reduced-biased exponential averaging layer to the attention mechanism. Based on the MEGA method of \cite{ma2022mega}, reduced-biased Hierarchical EMA is introduced as improved model. This improved architecture is shown in the \Cref{fig:rhema-structure}.

RHEMA uses GAU\cite{hua2022transformer} to share feature dimensions. First, pass it to the EMA layer at \Cref{eq:xema}, then use the SiLU self-gated activation function to linearly project to z dimension, and add nonlinear features.

\begin{equation}
    X_{EMA}=EMA(X_{in})
    \label{eq:xema}
\end{equation}

\begin{equation}
    Z=F_{SiLU}(X_{EMA}W_z+b_z) + X_{in}
    \label{eq:zsilu}
\end{equation}


With the advantage of SiLU, the operation here can compress the originally redundant d-dim input into a more compact z-dim representation while retaining the context information. \Cref{eq:silu} represents the specific calculation process of SiLU.

\begin{equation}
    F_{SiLU}=\sigma(x)+x \cdot \sigma(x) \cdot (1-\sigma(x))
    \label{eq:silu}
\end{equation}


Then, through the output of \Cref{eq:zsilu}, \textbf{Q}uery in \Cref{eq:emaq}, \textbf{K}ey in \Cref{eq:emak} and \textbf{V}alue in \Cref{eq:emav} is calculated. Among them, the bias terms ($\mu_q$, $\mu_k$ and $b_v$) add additional degrees of freedom to the attention mechanism, which allows the query and key to adjust their respective baseline values based on the shared representation $Z$ to make them more suitable for broader tasks. And $v$ refers to the extension dim of value series.

\begin{equation}
    Q = \mathcal{K}_q \odot Z + \mu_q
    \label{eq:emaq}
\end{equation}

\begin{equation}
    K = \mathcal{K}_k \odot Z + \mu_k
    \label{eq:emak}
\end{equation}

\begin{equation}
    V= F_{SiLU}(X_{in}W_b+b_v)
    \label{eq:emav}
\end{equation}


Next, the result of the key-value pair query can be expressed as \Cref{eq:query}.

\begin{equation}
    O=f_{S}(\frac{QK^T}{\daleth }+b_{rel})V
    \label{eq:query}
\end{equation}
where $f_{S}(x)$ means single headed attention, and the bias term of $b_{rel}$ can be optimally solved by attention. $\daleth$ represents the scaling factor, which reduces the digits of large queries.


by simulating the gating mechanism of GRU, activation output in \Cref{eq:yhat} is introduced. It affects the influence of each cell on the global affection.

\begin{equation}
    \hat{Y}=F_{SiLU}(X_{SiLU}W_h+(\gamma \odot O) U_h +b_h)
    \label{eq:yhat}
\end{equation}
Where the reset gate $\gamma$ controls whether the weight matrix $U_h$ affects the final output.

Finally, the output is represented as \Cref{eq:opt}.

\begin{equation}
    Y=\Phi \odot \hat{Y_{s}} + (1-\Phi ) \cdot X_{in}
    \label{eq:opt}
\end{equation}
where $\Phi$ represents the update gate, smoothing the result by Sigmoid activation function.

\begin{figure}[!htbp]
    \centering
    \includegraphics[width=0.5\linewidth]{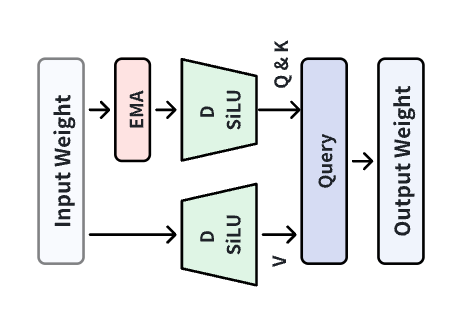}
    \caption{The architecture of single-head attention.}
    \label{fig:single-attention}
\end{figure}

The single-head attention calculation method mentioned in \Cref{eq:opt} can be expressed as \Cref{fig:single-attention}.

The output $Y$ combines the candidate activation state $\hat{Y}$ and the original input $X_{in}$, and the update gate $\Phi$ controls the information fusion ratio between the two parameters. When calculating the candidate activation state $\hat{H}$, the exponential average attention mechanism uses the Laplace attention function in \Cref{eq:laplace-origin} as the core component. Based on the concept of reduced-bias, it can be improved the reduced-bias Laplace function as \Cref{eq:laplace-fixed} to adapt to the introduction of deviation inside.

\begin{equation}
F_{la}(x; \mu, \sigma) = \frac{ 1 + \text{erf}\left(\frac{x - \mu}{\sigma \sqrt{2}}\right)}{2}
\label{eq:laplace-origin}
\end{equation}

\begin{equation}
F_{rla}(x; \mu, \sigma) = Norm(\frac{ 1 + \text{erf}\left(\frac{x - \mu}{\sigma \sqrt{2}}\right)}{2}+X_{in})
\label{eq:laplace-fixed}
\end{equation}

The Laplace attention function maps the attention score of $QK^T$ to a smooth range by dynamically adjusting the parameters $\mu$ and $\sigma$, providing a more stable gradient and a more flexible attention distribution. This smoothing range is mapped to (0, 1) using the erf function, which improves the global context modeling ability of the model, and ultimately improves the expression effect and task performance of $Y$.

\subsection{Reduced-bias Module}


Compared with the traditional residual connection layer, proposed work have improved on the basis of the classic residual network structure (Residual Network)\cite{he2016deep}, and added a dynamic adjustment mechanism for the network weights before and after the residual connection.

Specifically, the proposed Reduced-bias module fits the sensitivity and influence of different weight parameters on downstream gradient updates by constructing an adaptive weight adjustment strategy during the forward propagation process. In the verification feedback stage, the module can select the optimal parameter configuration through dynamic optimization, thereby enhancing the gradient transfer efficiency and learning ability of the model. In addition, this improvement can effectively alleviate the information transfer bias problem caused by fixed weights in traditional residual connections, and provide stronger robustness and flexibility for deep feature extraction and learning of the model.


Classic resnet can be expressed as \Cref{eq:nr}.
\begin{equation}
    \mathbf{y}_l=\mathcal{F}(\mathbf{x}_l,\mathbf{W}_l)+\mathbf{x}_l
    \label{eq:nr}
\end{equation}
where $\mathbf{x}_l$ is the input of the $l$-th layer module. $\mathcal{F}(\cdot)$ represents the operation of the network in this layer. $\mathbf{W}_l$ is a parameter of the $l$-th module.


Among \Cref{eq:nr}, dynamic adaptation mechanism parameters $\alpha$ and $\beta$ is introduced. The new network can be expressed as \Cref{eq:rb}.

\begin{equation}
    \mathbf{y}_l=\alpha_l\cdot\mathcal{F}(\mathbf{x}_l,\mathbf{W}_l)+\beta_l\cdot\mathbf{x}_l
    \label{eq:rb}
\end{equation}
where $\alpha$ controls the influence of the residual term, and $\beta$ controls the influence of the input direct connection term. Besides, $l \in L$, $L$ represents all the layer connected by Reduced-biased module.

\begin{equation}
    t_l = \sigma\left( \frac{\partial \mathcal{L}}{\partial \mathcal{F}(\mathbf{x}_l, \mathbf{W}_l)} \cdot \mathbf{W}_{t} + b_{t} \right)
    \label{eq:al}
\end{equation}
where $\mathcal{L}$ is the loss function, $\mathbf{W}_{\alpha}$ and $\mathbf{W}_{\beta}$ are learnable weight matrices, $b_{\alpha}$ and $b_{\beta}$ are bias terms, and $\sigma(\cdot)$ is the activation function.


By deriving \Cref{eq:al}, $\mathbf{y}_l$ can be got by \Cref{eq:redbia}.

\begin{equation}
\begin{aligned}
\mathbf{y}_l = & \, \sigma\left( \frac{\partial \mathcal{L}}{\partial \mathcal{F}(\mathbf{x}_l, \mathbf{W}_l)} \cdot \mathbf{W}_{\alpha} + b_{\alpha} \right) 
\cdot \mathcal{F}(\mathbf{x}_l, \mathbf{W}_l) \\
& + \sigma\left( \frac{\partial \mathcal{L}}{\partial \mathbf{x}_l} \cdot \mathbf{W}_{\beta} + b_{\beta} \right) 
\cdot \mathbf{x}_l
\end{aligned}
\label{eq:redbia}
\end{equation}

Using optimization of the parameters of this layer, the contribution of different paths in the residual connection to the final output can be dynamically adjusted, thus achieving adaptive learning and weight optimization.

\subsection{Robustly Optimized BERT}




RoBERTa (Robustly Optimized BERT Approach)\cite{cui2020revisiting} is an improved version of BERT proposed by Facebook AI. It greatly improves the performance of the model by optimizing the training strategy and hyperparameters. Compared with BERT, RoBERTa has been enhanced and improved in many aspects. First of all, RoBERTa uses a larger data set for training, and the total amount of data is more than ten times larger than BERT. This significantly improves the language representation capabilities of the model.

In addition, RoBERTa removes the Next Sentence Prediction task in BERT and only retains the training target based on Masked Language Model. Research shows that removing NSP does not reduce model performance, but reduces unnecessary constraints, thereby improving task performance. At the same time, RoBERTa introduces a dynamic masking strategy instead of BERT’s static masking. This means that the mask is re-randomly generated for the input text each time, thereby enhancing the diversity of training and the robustness of the model.

Through the above improvements, RoBERTa has achieved excellent performance on multiple natural language understanding tasks, including benchmarks such as GLUE, RACE, and SQuAD. Compared with BERT, RoBERTa's more powerful expression ability and flexible training strategy make it a pre-trained language model with outstanding performance in many downstream tasks.

\section{Experiments}
\label{sec:experiments}

\subsection{Datasets Description}


Three NER benchmark datasets are collected to verify the performance of proposed model. They are from Weibo\cite{peng2015named}, MSRA\cite{zhang2006word}, and Resume\cite{zhang2018chinese}.

The specific properties of the original datasets are shown in the \Cref{tab:datasets}. It should be noted that since MSRA does not have a validation set by default, all experiments directly use the test-set as the valid-set. For the other two datasets, a split rate of 70\%:15\%:15\% is used to evaluate.

\begin{table}[!htbp]
\centering
\caption{Properties of the three datasets used.}
\begin{tabular*}{\linewidth}{@{\extracolsep{\fill}} lccc @{}}
\toprule
\textbf{Datasets} & \textbf{Weibo} & \textbf{MSRA} & \textbf{Resume} \\ 
\midrule
Class          & 8         & 3        & 8          \\
Train          & 1,350     & 46,364   & 3,821      \\
Test           & 270       & 4,365    & 477        \\
Valid          & 269       & 4,365    & 463        \\
Avg length     & 54.61     & 46.80    & 32.47      \\
Max length     & 175       & 581      & 178        \\
Min length     & 7         & 5        & 3          \\ 
\bottomrule
\end{tabular*}
\label{tab:datasets}
\end{table}

\subsection{Experiment Settings}



To verify the versatility and effectiveness of the proposed model in the NER task, BiLSTM-CRF is used BiLSTM-CRF as the baseline model. On this basis, comparative experiments and ablation experiments are designed. In addition, it is also compared the impact of different pre-training heads on model indicators.

Experimental environment is based on the Ubuntu 22.04LTS operating system, using the NVIDIA RTX4090 for inference and computing, and the cuda version is 12.4, with Python 3.9 environment. During the training process, Adam is used to optimize the learning rate. The number of training epochs is set to 100 times with early stop callback.

\subsection{Evaluation Metrics}


Proposed work uses precision (P), Recall (R), and F-Score ($F_1$) to evaluate the model.

The negative log-likelihood function, shown as \Cref{eq:nll} is considered as the loss function to quantitatively evaluate the performance of the model.

\begin{equation}
\text{NLL} = -\sum_{i=1}^{N} \sum_{c=1}^{C} y_{i,c} \log(\hat{y}_{i,c})
\label{eq:nll}
\end{equation}
where $y_{i,c}$ is the binary indicator (0 or 1) of whether class label $c$ is the correct classification for sample $i$, and $\hat{y}_{i,c}$ is the predicted probability of sample $i$ belonging to class $c$.

\subsection{Experimental Results}

\subsubsection{Comparison Studies}
The experimental results on MSRA, Resume and Weibo for comparison studies are shown in the several tables.

\begin{table}[!htbp]
\centering
\caption{Comparison Experimental results on MSRA.}
\begin{tabular*}{\linewidth}{@{\extracolsep{\fill}} llccc @{}}
\toprule
\multirow{2}{*}{\textbf{Models}} & \multirow{2}{*}{\textbf{Description}} & \multicolumn{3}{c}{\textbf{MSRA}} \\ 
\cmidrule(lr){3-5}
 &  & \textbf{P} & \textbf{R} & \textbf{F1} \\ 
\midrule
BiLSTM+CRF &  & 84.65 & 80.68 & 82.62 \\
Lattice LSTM & Zhang 2018~\cite{zhang2018chinese} & 93.57 & 92.79 & 93.18 \\
AS & Cao 2018~\cite{cao2018adversarial} & 91.73 & 89.58 & 90.64 \\
TextCNN & Jia 2018~\cite{jia2018chinese} & 91.63 & 90.56 & 91.09 \\
DEM & Zhang 2019~\cite{zhang2019chinese} & 90.59 & 91.15 & 90.87 \\
MFE & Li 2021~\cite{li2021mfe} & 90.47 & 84.70 & 87.49 \\
MSFM & Zhang 2022~\cite{zhang2022chinese} & 86.90 & 85.46 & 86.17 \\
DAE & Liu 2024~\cite{liu2024dae} & 92.73 & 90.15 & 91.42 \\
\textbf{HREB} & \textbf{(Ours)} & \textbf{94.58} & \textbf{93.98} & \textbf{94.28} \\ 
\bottomrule
\end{tabular*}
\label{tab:comp-msra}
\end{table}

Compared with multiple existing methods on MSRA dataset by \Cref{tab:comp-msra}, the classic Lattice LSTM model achieves an F1 score of 93.18, demonstrating its effectiveness by combining character and word-level information. However, proposed method surpasses this benchmark with an F1 score of 94.28, demonstrating stronger semantic modeling capabilities. Compared with the 82.62 F1 score of BiLSTM+CRF as the baseline model, proposed structure shows significant advantages in deep feature extraction and context information utilization.

In addition, compared with DEM and DAE in recent years, proposed model has significantly improved both precision and recall, especially showing stronger generalization ability in complex entity recognition tasks. Comprehensive results show that proposed method achieves state-of-the-art performance on MSRA dataset.

\begin{table}[!htbp]
\centering
\caption{Comparison Experimental results at Resume.}
\begin{tabular*}{\linewidth}{@{\extracolsep{\fill}} llccc @{}}
\toprule
\multirow{2}{*}{\textbf{Models}} & \multirow{2}{*}{\textbf{Description}} & \multicolumn{3}{c}{\textbf{Resume}} \\ 
\cmidrule(lr){3-5}
 &  & \textbf{P} & \textbf{R} & \textbf{F1} \\ 
\midrule
BiLSTM+CRF & & 94.79 & 93.68 & 94.23 \\
Lattice LSTM & Zhang 2018~\cite{zhang2018chinese} & 94.81 & 94.11 & 94.46 \\
LLPA & Gan 2021~\cite{gan2021chinese} & 96.36 & 94.88 & 95.62 \\
MSFM & Zhang 2022~\cite{zhang2022chinese} & 96.08 & 94.79 & 95.43 \\
MFT & Han 2022~\cite{han2022multi} & 96.05 & 95.52 & 95.78 \\
DAE & Liu 2024~\cite{liu2024dae} & \textbf{96.92} & 95.18 & 96.04 \\
\textbf{HREB} & \textbf{(Ours)} & 95.53 & \textbf{96.61} & \textbf{96.07} \\ 
\bottomrule
\end{tabular*}
\label{tab:comp-resume}
\end{table}

Although proposed model has achieved meaningful recall (96.61) and F1 score (96.07) on the Resume dataset, detailed in \Cref{tab:comp-resume}. There are still some shortcomings that deserve attention. First, compared with the DAE model, the precision is slightly lower (95.53 vs. 96.92), indicating that there is space for improvement in the prediction accuracy of specific entities.

\begin{table}[!htbp]
\centering
\caption{Comparison Experimental results at Weibo.}
\begin{tabular*}{\linewidth}{@{\extracolsep{\fill}} llccc @{}}
\toprule
\multirow{2}{*}{\textbf{Models}} & \multirow{2}{*}{\textbf{Description}} & \multicolumn{3}{c}{\textbf{Weibo}} \\ 
\cmidrule(lr){3-5}
 &  & \textbf{P} & \textbf{R} & \textbf{F1} \\ 
\midrule
BiLSTM+CRF & & 57.84 & 49.08 & 53.10 \\
Lattice LSTM & Zhang 2018~\cite{zhang2018chinese} & 53.04 & 62.25 & 58.79 \\
Glyce & Meng 2019~\cite{meng2019glyce} & 67.68 & 67.71 & 67.60 \\
FLAT & Li 2020~\cite{li2020flat} & - & - & 63.42 \\
FLAT+BERT & Li 2020~\cite{li2020flat} & - & - & 68.55 \\
MSFM & Zhang 2022~\cite{zhang2022chinese} & 60.75 & 51.83 & 55.94 \\
DAE & Liu 2024~\cite{liu2024dae} & \textbf{69.68} & 48.89 & 57.45 \\ 
\textbf{HREB} & \textbf{(Ours)} & 66.57 & \textbf{70.76} & \textbf{68.60} \\ 
\bottomrule
\end{tabular*}
\label{tab:comp-weibo}
\end{table}

\begin{figure*}
    \centering
    \includegraphics[width=\linewidth]{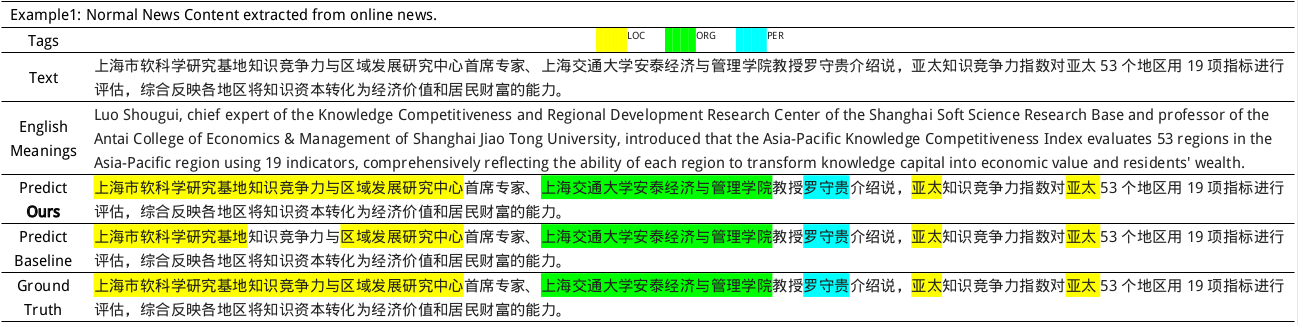}
    \caption{Based on the web search results, a piece of news was randomly selected as a test sample, which contains the extraction of names of people, places and organizations.}
    \label{fig:example1}
\end{figure*}

Proposed method achieves an F1 score of 68.60, slightly higher than FLAT+BERT (68.55) and close to Glyce (67.60), demonstrating robustness in complex text processing on Weibo, shown as \Cref{tab:comp-weibo}. Furthermore, proposed method performs well in recall (70.76), indicating its advantage in identifying more potential entities. However, compared with the precision of DAE (69.68), proposed method still has a certain gap in accuracy, which suggests that proposed model still has room for improvement in accurate prediction when dealing with noisy social media data.

\subsubsection{Ablation Studies}
Ablation Study in \Cref{tab:ablation-msra} evaluates the proposed structure through two experiments: (1) module removal to assess synergy, and (2) BERT replacement/removal to analyze word embedding impacts.

\paragraph{Comparison between the base model and different BERT models} The experimental results of comparing the base model (S.N.1) with the models that introduce different BERT variants (S.N.2 and S.N.3) show that RoBERTa outperforms BERT in terms of vocabulary vectorization.

\paragraph{Comparison between naive attention and hierarchical EMA} When comparing the naive attention model (S.N.4) with the model based on hierarchical exponential moving average (Hierarchical EMA) (S.N.5), the experimental results show that the hierarchical attention mechanism can more effectively extract the weight distribution of key information under the constraint of exponential gain weights.

\paragraph{Comparison between static and dynamic reduced-bias models} By comparing the immutable Reduced-bias model (S.N.4) with the self-learning Reduced-bias model (S.N.5), the results verify that the dynamically changing coefficients can be adaptively adjusted according to the input text features, thereby improving the model's ability to express features.

\paragraph{Comparison of the proposed model with the naive BERT model} In the comparison of the final model (S.N.8) with the ordinary BERT model (S.N.7), the experimental results show that even with the introduction of a rich feature extraction module, RoBERTa can still efficiently capture the deep features of the text and show a superior feature learning ability.

\begin{table}[!htbp]
\centering
\caption{Ablation Studies on MSRA}
\begin{tabular*}{\linewidth}{@{\extracolsep{\fill}} cccccccc @{}}
\toprule
\multirow{2}{*}{\textbf{S.N.}} & \multicolumn{4}{c}{\textbf{Module}} & \multicolumn{3}{c}{\textbf{Metric}} \\
\cmidrule(r){2-5} \cmidrule(l){6-8}
& \textbf{A} & \textbf{H} & \textbf{R} & \textbf{BERT} & \textbf{P} & \textbf{R} & \textbf{F1} \\ 
\midrule
\multicolumn{1}{c|}{1} & & & & \multicolumn{1}{c|}{} & \multicolumn{1}{c}{84.65} & \multicolumn{1}{c}{80.68} & \multicolumn{1}{c}{82.62} \\
\multicolumn{1}{c|}{2} & & & & \multicolumn{1}{c|}{BERT-base} & \multicolumn{1}{c}{91.02} & \multicolumn{1}{c}{92.09} & \multicolumn{1}{c}{91.55} \\
\multicolumn{1}{c|}{3} & & & & \multicolumn{1}{c|}{RoBERTa} & \multicolumn{1}{c}{92.21} & \multicolumn{1}{c}{92.57} & \multicolumn{1}{c}{92.38} \\
\multicolumn{1}{c|}{4} & $\surd$ & & & \multicolumn{1}{c|}{} & \multicolumn{1}{c}{87.42} & \multicolumn{1}{c}{85.78} & \multicolumn{1}{c}{86.60} \\
\multicolumn{1}{c|}{5} & & $\surd$ & & \multicolumn{1}{c|}{} & \multicolumn{1}{c}{91.35} & \multicolumn{1}{c}{90.62} & \multicolumn{1}{c}{90.98} \\
\multicolumn{1}{c|}{6} & & $\surd$ & $\surd$ & \multicolumn{1}{c|}{} & \multicolumn{1}{c}{92.92} & \multicolumn{1}{c}{93.09} & \multicolumn{1}{c}{93.01} \\
\multicolumn{1}{c|}{7} & & $\surd$ & $\surd$ & \multicolumn{1}{c|}{BERT-base} & \multicolumn{1}{c}{93.55} & \multicolumn{1}{c}{93.88} & \multicolumn{1}{c}{93.71} \\ 
\midrule
\multicolumn{1}{c|}{\textbf{8 (Ours)}} & & $\surd$ & $\surd$ & \multicolumn{1}{c|}{RoBERTa} & \multicolumn{1}{c}{\textbf{94.58}} & \multicolumn{1}{c}{\textbf{93.98}} & \multicolumn{1}{c}{\textbf{94.28}} \\ 
\bottomrule
\end{tabular*}
\label{tab:ablation-msra}
\end{table}

\subsection{Parameter Selection}

In the design of the attention module based on the EMA (Exponential Moving Average) mechanism, the number of attention heads is considered to be one of the important factors affecting the model performance. In order to determine the optimal configuration of attention heads, different numbers of attention heads (16, 32, 64, 128) is set in the experiment and evaluated the performance of the model on the validation set. The experimental results show that when the number of attention heads is set to 64, the model achieves the best performance in performance indicators such as precision, recall, and F1 score.

\begin{table}[!htbp]
\centering
\caption{Effects of different parameters on performance in RHEMA.}
\begin{tabular*}{\linewidth}{@{\extracolsep{\fill}} cccccc @{}}
\toprule
\multicolumn{2}{c}{\multirow{2}{*}{\textbf{EMA head}}} & \multirow{2}{*}{\textbf{Metric}} & \multicolumn{3}{c}{\textbf{Dataset}} \\ 
\cmidrule(l){4-6} 
\multicolumn{2}{c}{} & & \textbf{MSRA} & \textbf{Weibo} & \textbf{Resume} \\ 
\midrule
\multicolumn{1}{c|}{\multirow{15}{*}{n\_head}} & \multirow{3}{*}{8} & P & 92.42 & 65.96 & 92.24 \\
\multicolumn{1}{c|}{} & & R & 93.44 & 68.24 & 94.96 \\
\multicolumn{1}{c|}{} & & F1 & 92.93 & 67.08 & 93.58 \\ 
\cmidrule(lr){2-6} 
\multicolumn{1}{c|}{} & \multirow{3}{*}{16} & P & 93.73 & 65.90 & 94.08 \\
\multicolumn{1}{c|}{} & & R & 94.59 & 71.70 & 95.31 \\
\multicolumn{1}{c|}{} & & F1 & 94.16 & 68.67 & 94.69 \\ 
\cmidrule(lr){2-6} 
\multicolumn{1}{c|}{} & \multirow{3}{*}{32} & P & 93.89 & 66.27 & 94.52 \\
\multicolumn{1}{c|}{} & & R & 94.86 & 70.44 & 95.92 \\
\multicolumn{1}{c|}{} & & F1 & 94.37 & 68.29 & 95.21 \\ 
\cmidrule(lr){2-6} 
\multicolumn{1}{c|}{} & \multirow{3}{*}{64} & P & \textbf{94.58} & \textbf{66.57} & \textbf{95.53} \\
\multicolumn{1}{c|}{} & & R & \textbf{93.98} & \textbf{70.76} & \textbf{96.61} \\
\multicolumn{1}{c|}{} & & F1 & \textbf{94.28} & \textbf{68.60} & \textbf{95.07} \\ 
\cmidrule(lr){2-6} 
\multicolumn{1}{c|}{} & \multirow{3}{*}{128} & P & 93.92 & 64.77 & 95.16 \\
\multicolumn{1}{c|}{} & & R & 94.50 & 71.70 & 95.74 \\
\multicolumn{1}{c|}{} & & F1 & 94.21 & 68.06 & 95.45 \\ 
\bottomrule
\end{tabular*}
\label{tab:param-rhema}
\end{table}

As seen from the \Cref{tab:param-rhema}, too few attention heads with 16 or 32, may cause the model to be unable to fully capture feature representations, thereby limiting the performance of the model. When the number of attention heads increases to 64, the performance of the model reaches its peak. However, when the number of attention heads is further increased to 128), the performance of the model does not continue to improve, but instead shows a slight decline. This phenomenon may be due to the fact that too many attention heads increase the complexity of the model, thereby introducing the risk of overfitting.

\subsection{Case Analysis}
\paragraph{Real Example}Select text data on the Internet for analysis.

In Example-1 of \Cref{fig:example1}, the performance of the model trained on MSRA dataset is verified using normal press releases. The results show that proposed model can effectively extract entity information for regular text. For the baseline model, it recognizes the text of the school-college combination as two different entities. When considering ground truth, it is generally believed that school+college is a complete entity.

\begin{figure*}[!htbp]
    \centering
    \includegraphics[width=0.8\linewidth]{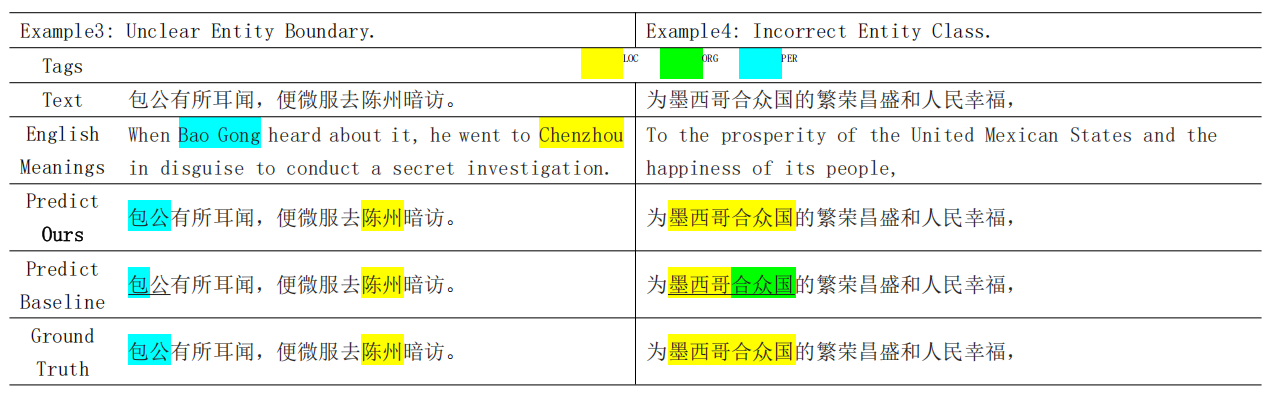}
    \caption{Cases in the MSRA validation set are selected for analysis.}
    \label{fig:example34}
\end{figure*}

\begin{figure}
    \centering
    \includegraphics[width=\linewidth]{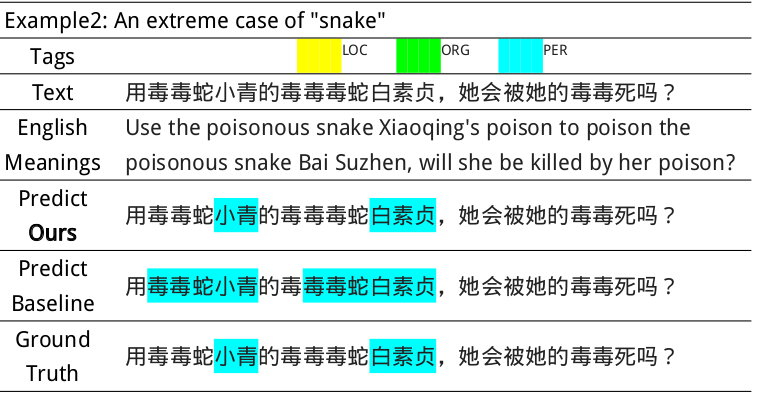}
    \caption{In the classic Chinese story “The Legend of the White Snake”, Bai Suzhen and Xiaoqing are female protagonists who were transformed from poisonous snakes into human form. This case involves the extraction of obfuscated names.}
    \label{fig:example2}
\end{figure}

In Example-2 of \Cref{fig:example2}, an extreme scenario was constructed. The same Chinese character was expressed as a noun, a verb, and an adjective simultaneously. When these were combined into a sentence, it was verified whether the model could output correctly in extreme cases. The experimental results indicate that the two protagonists could be effectively identified by the proposed model. However, for the baseline model, although the two protagonists were covered, the entity boundary was found to be unclear.

\paragraph{Dataset Example}Select text data on the datasets for analysis.

The analysis presented in \Cref{fig:example34} addresses two primary challenges in entity recognition, as outlined in the introduction: 1) ambiguity in boundary identification (example-3); 2) errors in entity recognition (example-4). A comparative evaluation demonstrates that the proposed model effectively resolves these issues in the selected examples, exhibiting superior performance relative to the baseline model.

\subsection{Analysis and Discussion}

Through comparative and ablation studies, key findings on model performance are summarized:

1) Proposed model adapts to diverse semantic information across scene tasks, enhancing CNER robustness. As shown in \Cref{tab:comp-msra,tab:comp-weibo,tab:comp-resume}, proposed method achieves top performance in regression and F1 scores. Ablation studies in \Cref{tab:ablation-msra} confirm that hierarchical exponential attention and reduced-bias modules expand data depth, excelling in conventional tasks in \Cref{fig:example1} and domain tasks in \Cref{fig:example2,fig:example34}.

2) The Hierarchical Reduced-bias EMA architecture enables precise CNER feature extraction. Inspired by DAE \cite{liu2024dae}, the model extends to sequence tasks (e.g., text classification, emotion recognition) using BERT-based vectorization.

3) Increasing attention heads raises parameters, prolonging training/inference time. In order to balanceto balance performance and efficiency, 64 EMA heads were selected.

\section{Conclusion}
\label{sec:conclusion}

Proposed study focus on CNER model that combines RoBERTa, reduced-biased Hierarchical EMA, BiLSTM, and CRF, which conducts extensive experimental verification on multiple Chinese datasets. Experimental results show that proposed method has achieved leading performance on MSRA, Resume, and Weibo datasets, demonstrating strong adaptability to diverse language scenarios. Despite the excellent experimental results, there are still some directions worth exploring. In the future, stronger feature fusion strategies, more robust pre-trained models, and regularization techniques that adapt to diverse scenarios will be explored to further improve model performance and generalization capabilities.

\section*{Acknowledgments}
This research was funded by Programme MARS (Programme of Maritime AI Research in Singapore) with funding grant number SMI-2022-MTP-06 by Singapore Maritime Institute (SMI).


\end{document}